\documentclass[lettersize,journal]{IEEEtran}
\usepackage{amsmath,amsfonts}
\usepackage{algorithmic}
\usepackage{algorithm}
\usepackage{array}
\usepackage[caption=false,font=normalsize,labelfont=sf,textfont=sf]{subfig}
\usepackage{textcomp}
\usepackage{stfloats}
\usepackage{url}
\usepackage{verbatim}
\usepackage{graphicx}
\usepackage{cite}
\usepackage{amssymb}
\makeatletter

\newcommand{\Rmnum}[1]{\expandafter\@slowromancap\romannumeral #1@}
\makeatother

\begin{document}

\title{Aerial-NeRF: Adaptive Spatial Partitioning and Sampling for Large-Scale Aerial Rendering}


\author{
Xiaohan~Zhang,
Yukui~Qiu,
Zhenyu~Sun,
and~Qi~Liu, \IEEEmembership{IEEE Senior Member}
\thanks{The authors are with the School of Future Technology, South China University of Technology, Guangzhou 511442, China.
E-mail: \{ftxiaohanzhang, 202162311356, 202264690427\}@mail.scut.edu.cn, drliuqi@scut.edu.cn (Corresponding author: Qi Liu).}%
} 

\markboth{Journal of \LaTeX\ Class Files,~Vol.~14, No.~8, August~2021}%
{Shell \MakeLowercase{\textit{et al.}}: A Sample Article Using IEEEtran.cls for IEEE Journals}


\maketitle

\begin{abstract}

  Recent progress in large-scale scene rendering has yielded Neural Radiance Fields (NeRF)-based models with an impressive ability to synthesize scenes across 
  small objects and indoor scenes. Nevertheless, extending this idea to
  large-scale aerial rendering poses two critical problems. Firstly, a single NeRF cannot render the entire scene
  with high-precision for complex large-scale aerial datasets since the sampling range along each view ray is insufficient to cover buildings adequately. 
  Secondly, traditional NeRFs are infeasible to train on one GPU to enable interactive fly-throughs for modeling massive images.
  Instead, existing methods typically separate the whole scene into multiple regions and train a NeRF on each region,
  which are unaccustomed to different flight trajectories and difficult to achieve fast rendering.
  To that end, we propose Aerial-NeRF with three innovative modifications for jointly adapting NeRF in large-scale aerial rendering: 
  (1) Designing an adaptive spatial partitioning and selection method based on drones' poses to adapt different flight trajectories;
  (2) Using similarity of poses instead of (expert) network for rendering speedup to
  determine which region a new viewpoint belongs to; (3) Developing an adaptive sampling approach for rendering performance improvement
  to cover the entire buildings at different heights.
  Extensive experiments have conducted to verify the effectiveness and efficiency of Aerial-NeRF, and new state-of-the-art results have been achieved on two public 
  large-scale aerial datasets and presented SCUTic dataset.
  Note that our model allows us to perform rendering over 4
  times as fast as compared to multiple competitors. Our dataset, code, and model are publicly available at https://drliuqi.github.io/.

\end{abstract}


\section{Introduction}

\IEEEPARstart{N}{eural} Radiance Fields (NeRF) \cite{mildenhall2021nerf} synthesizes highly realistic 3D scenes from limited observations
due to its implicit scene representation. NeRF has pervasive for rendering small objects and indoor scenes \cite{barron2021mip} \cite{muller2022instant}
\cite{chen2022tensorf} \cite{yang2022recursive} in
illumination, reflections, and texture-less areas, which outperforms previous methods based on mesh \cite{nimier2019mitsuba}
\cite{liu2019soft} \cite{li2018differentiable} \cite{buehler2023unstructured} and voxel \cite{seitz1999photorealistic} \cite{lombardi2019neural}
\cite{sitzmann2019deepvoxels} \cite{mildenhall2019local} with ever growing popularitiess \cite{deng2022fov} \cite{zhong2023vq} \cite{zhang2024text2nerf} 
\cite{wang2022nerfcap}. 

NeRF has also evolved from rendering small objects to more complex scenes. 
Such modeling can enable a variety of practical applications, including
autonomous vehicle simulation \cite{li2019aads} \cite{ost2021neural} \cite{yang2020surfelgan}, aerial surveying \cite{bozcan2020air} \cite{du2018unmanned}, 
and embodied AI \cite{morad2021embodied} \cite{truong2021bi}.
Considering how to sample on objects at infinity, NeRF++ \cite{zhang2020nerf++} proposes to compress unbounded scenes into bounded regions 
enhancing the rendering effect of distant scenes. 
Mip-NeRF \cite{barron2021mip} replaces NeRF rays with view frustums and utilizes the structural information to achieve more accurate rendering results.
This method can adaptively encode the inputs, via using low-frequency positional encoding in sparse sampling areas, and high-frequency positional encoding 
in dense sampling areas to achieve an anti-aliasing effect.
NeRF-W \cite{martin2021nerf} proposes appearance and transient embedding to handle changes in illumination and dynamic objects for the rendering 
quality enhancement of outdoor scenes.
Block-NeRF \cite{tancik2022block} divides the street scene into multiple areas, and trains a NeRF separately in each area, enhancing the rendering accuracy of 
texture details.
However, the deployment of NeRF-based models on large-scale aerial scenes is still impeded by two problems: (1) 
Using only a NeRF to render the 
whole scene can result in insufficient detail expression and excessive GPU memory consumption for high-resolution images in large-scale aerial datasets.
(2) As for the far distance between the camera and the buildings, it becomes necessary to design adaptive sampling algorithm for NeRF to cover the buildings.

\begin{figure}[!t]
  \centering
  \includegraphics[width=3.4in]{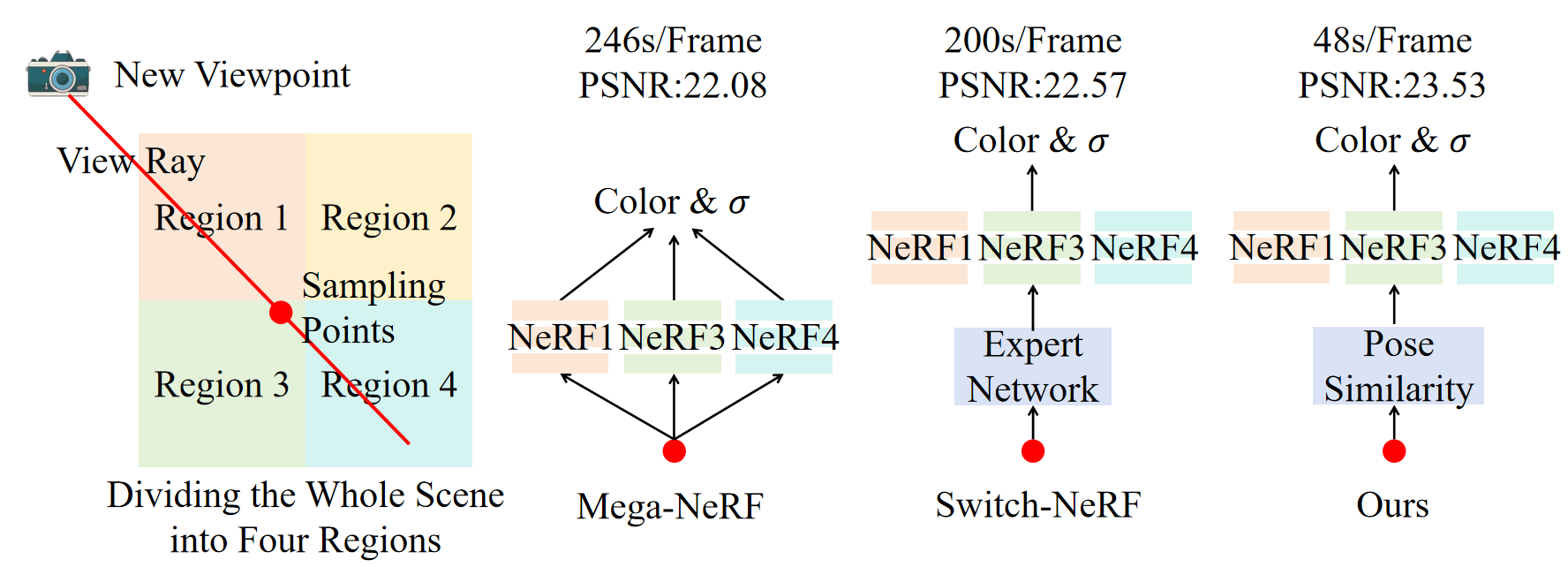}
  \caption{Comparison of different methods for rendering new viewpoints. 
  For a sampling point on the view ray, Mega-NeRF \cite{turki2022mega} uses NeRFs of all regions traversed by this ray to calculate the color and 
  density of this sampling point, resulting in a plodding rendering speed. Switch-NeRF \cite{zhenxing2022switch} applies an expert network to determine which region a sampling point belongs 
  to and applies the corresponding NeRF to calculate its color and density, thereby improving the rendering speed. Our method creatively utilizes existing camera 
  poses to match the region and the new viewpoint, which speeds up rendering.
  Moreover, ours is more robust for different aerial photography trajectories to achieve higher PSNR.}
  \label{pre_me}
  \end{figure}

Motivated by the Block-NeRF \cite{tancik2022block}, Mega-NeRF \cite{turki2022mega} evenly partitions the scene into 
multiple predefined regions to achieve arbitrarily 
large-scale scene rendering with only a single GPU, whereas it is unsuitable for uneven distributed drones. Switch-NeRF \cite{zhenxing2022switch} performs 
region partitioning by learning to achieve good rendering results. 
Nevertheless, Switch-NeRF requires to input all images at once, which takes up high GPU memory consumption non-amicable to limited memory resources when the scene size grows.
As shown in Fig. 1, both of them apply the (expert) network to determine 
the region which new perspective belongs to, which cost demanding inference time.
To address that, we design an efficient spatial partitioning and selection method, which clusters the poses of drones to partition the scene into multiple regions and 
selects cameras for each region based on boundary and similarity conditions.
The purpose of selecting cameras is to ensure that enough cameras can observe a certain region to train the region's NeRF.
To achieve sampling from unbounded space to bounded space, NeRF++ \cite{zhang2020nerf++} is developed to partition the unbounded space into a bounded 
foreground and an unbounded background,
where the foreground remains unaltered and the latter is compressed and transformed to a bounded space. 
However, the spatial gap between drones and the buildings
results in numerous sampling points distributed in the air, leading to a waste of 
sampling points and the GPU memory increase. Moreover, the sampling range can not cover buildings adequately, as shown in Fig. 2. 
To that end, we propose an innovative strategy for 
ensuring that the sampling range covers all buildings and the sampling points are almost distributed on the buildings
when cameras are at different heights.
We sample between the highest building and the surface of the Earth, so that the sampling range covers 
all buildings and most of the sampling points are distributed on buildings.  

\begin{figure}[!t]
  \centering
  \includegraphics[width=3.4in]{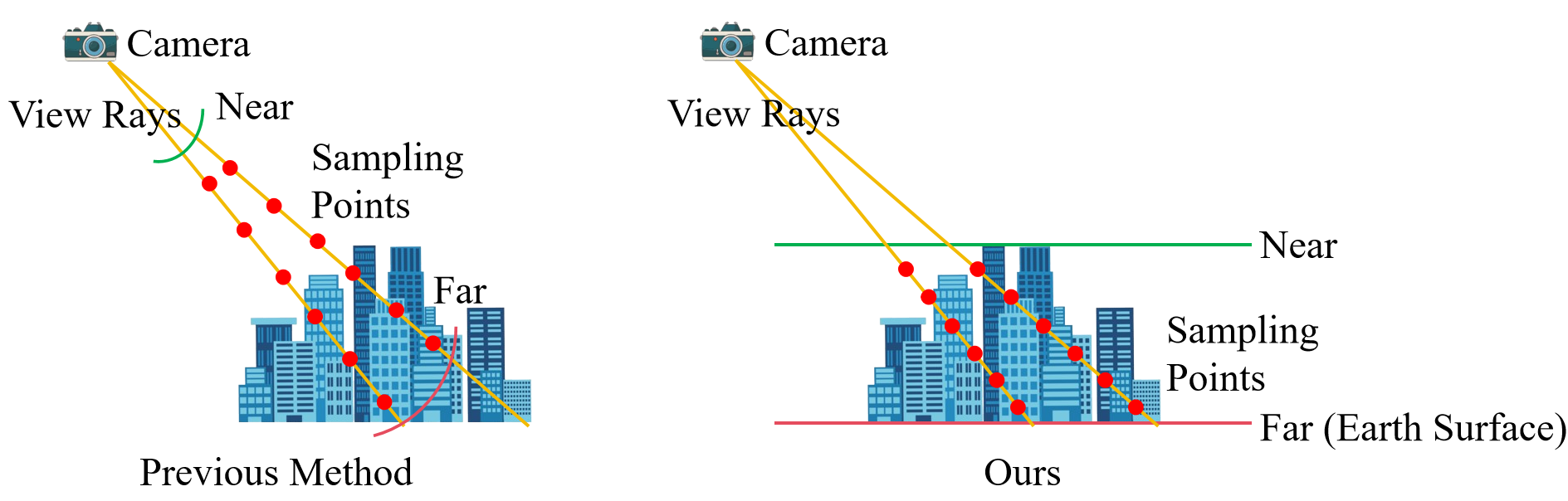}
  \caption{Comparison between different sampling approaches. "Near" represents sampling origin along each view ray, 
  and "Far" denotes the sampling end point. The previous sampling method sets the sampling range along each ray as a hyperparameter, 
  resulting in a significant waste of sampling points in the air and cannot cover the entire buildings in the sampling range. }
  \label{sample}
  \end{figure}

Our work makes notable contributions summarized as follows:

$\bullet$ For large-scale aerial rendering, we propose an adaptive spatial partitioning and selection approach based on the camera's pose (position and orientation of the camera). 
The proposed method outperforms existing large-scale aerial rendering models by a large margin on the rendering speed, almost at 4 times.
Besides, it is applicable to aerial datasets with 
diverse aerial photography trajectories. Additionally, under the appropriate number of divided regions, our method enables the rendering of arbitrarily large aerial scenes using a single GPU.

$\bullet$ We introduce a novel sampling strategy for aerial scenes, which enables to cover buildings by the sampling ranges from cameras at different heights.
In a broader comparison against SOTA models, our approach is substantially more efficient (only 1/4 used sampling points and 2 GB GPU memory saving) and 
compares favourably in terms of multiple commonly-used metrics.

$\bullet$ We present SCUTic, a novel aerial dataset for large-scale university campus scenes, which includes 5.86 GB high-resolution oblique photography images.
Unlike existing datasets, we collect data in a way that the camera trajectory is uneven, which can verify the robustness of rendering methods.

\section{Related Work}
Aerial datasets differ from other image datasets because there is significant space between cameras and the buildings of 
interest. Performing perspective rendering on large-scale aerial datasets is challenging, and this work is gradually 
gaining attention. We propose a novel spatial division and selection approach, along with a novel sampling strategy, achieving state-of-the-art 
rendering results. Additionally, we create a new drone aerial dataset using a novel aerial capture strategy distinct from previous approaches. 
This dataset is utilized to validate the robustness of rendering models. We consider the most closely related works below.

\subsection{NeRF for General Outdoor Scenes}
NeRF \cite{mildenhall2021nerf}  uses the MLP to map each sampling point's spatial position and view direction along view rays to color and density. 
By performing volume rendering integration along each view ray, the corresponding rendering image for a given camera view can be obtained. 
Due to the superiority of NeRF in view synthesis, many works improve its efficiency \cite{sun2022direct}, accuracy \cite{wang20224k} and apply 
it to 3D reconstruction tasks \cite{zhang2021ners} \cite{jiang2023nert}.

NeRF also has evolved from rendering small objects to more complex scenes. We primarily focus on the application of NeRF in large-scale outdoor scenes.
NeRF++ \cite{zhang2020nerf++} analyzes the reasons for the success of NeRF and proposes a method to compress unbounded scenes into bounded regions.
Mip-NeRF \cite{barron2021mip} proposes replacing NeRF rays with view frustums, utilizing the structural information to achieve more accurate rendering effects.
NeRF-W \cite{martin2021nerf} introduces algorithms to handle changes in illumination and dynamic objects, significantly enhancing the rendering quality of outdoor scenes.
Block-NeRF \cite{tancik2022block} suggests dividing large scenes into many regions and training a separate NeRF for each region. This approach allows for rendering scenes of 
arbitrary size. 

However, when rendering large-scale aerial scenes, the performance of these methods can not meet expectations.
In typical outdoor scene datasets, buildings far from the camera are considered background and not the main focus of attention.
In aerial datasets, buildings on the ground are far from the camera, but they are the scenes of interest for rendering. Therefore, sampling 
strategies and other related tactics need to be redesigned for aerial datasets.

\subsection{NeRF for Large-Scale Aerial Scenes}
This is a long-standing problem in computer vision \cite{agarwal2011building} \cite{fruh2004automated} \cite{li2008modeling} \cite{pollefeys2008detailed} \cite{schonberger2016structure} \cite{snavely2006photo} \cite{zhu2018very}.
For large-scale aerial scenes, it is essential to consider two key issues. The first issue is the sampling problem, determining how to ensure 
the sampling range covers objects on the ground. The second issue is region partitioning and selection because processing all images at once would require a 
high GPU memory. And the amount of data can exceed the expressive capacity of NeRF, resulting in blurry detail rendering.

Mega-NeRF \cite{turki2022mega} can render scenes of arbitrary size. This method uniformly divides the scene into several regions. Then, it crops each photo to retain only 
the pixels visible in that particular region. The sampling strategy of Mega-NeRF \cite{turki2022mega} is similar to NeRF++ \cite{zhang2020nerf++}. It involves sampling in both bounded regions 
and compressed unbounded regions. Eventually, the sampled points are fused to generate the color of the corresponding pixels. However, Mega-NeRF \cite{turki2022mega} sets 
the sampling range as a hyperparameter during sampling. This leads to the sampling range not covering objects and a significant number of 
sampled points in the air, causing an inaccurate rendering and a waste of sampling points. 
During rendering, Mega-NeRF \cite{turki2022mega} uses NeRFs of all regions that the view ray passes through to fuse the color and density of each sampling point on this ray,
leading to a slow inference speed. And Mega-NeRF \cite{turki2022mega} requires a specific aerial photography trajectory, where the camera positions should be 
distributed as uniformly as possible in the space. If the distribution of drones in space is uneven, this algorithm fails.

Switch-NeRF \cite{zhenxing2022switch} partitions the scene into several regions through a learning-based approach. 
This method requires inputting all the images and 
then learning the category of each pixel to partition the scene. As the scene size increases, this method requires computational resources to increase linearly. 
And the shape of the space divided by this method is unknown, adversely affecting the rendering results when the distribution of drones is uneven.
Moreover, this method shares the same sampling strategy as Mega-NeRF, leading to the need for numerous sampling points and situations 
where the sampling range can not cover objects.

In response, we propose a novel method, Aerial-NeRF, with solid efficiency and robustness for large-scale aerial datasets. Firstly, our spatial partitioning and selection
approach allows fast and accurate rendering. Secondly, our proposed sampling method enables sampling near objects, achieving high-quality rendering with 
minimal sample points. Moreover, our method can render aerial scenes of arbitrary size using only a single GPU. Finally, we introduce a new aerial dataset. 
The cameras' trajectory in our dataset differs from previous datasets, providing a good validation of the method's robustness.

\section{Method}

\begin{figure*}[!t]
\centering
\includegraphics[width=7in]{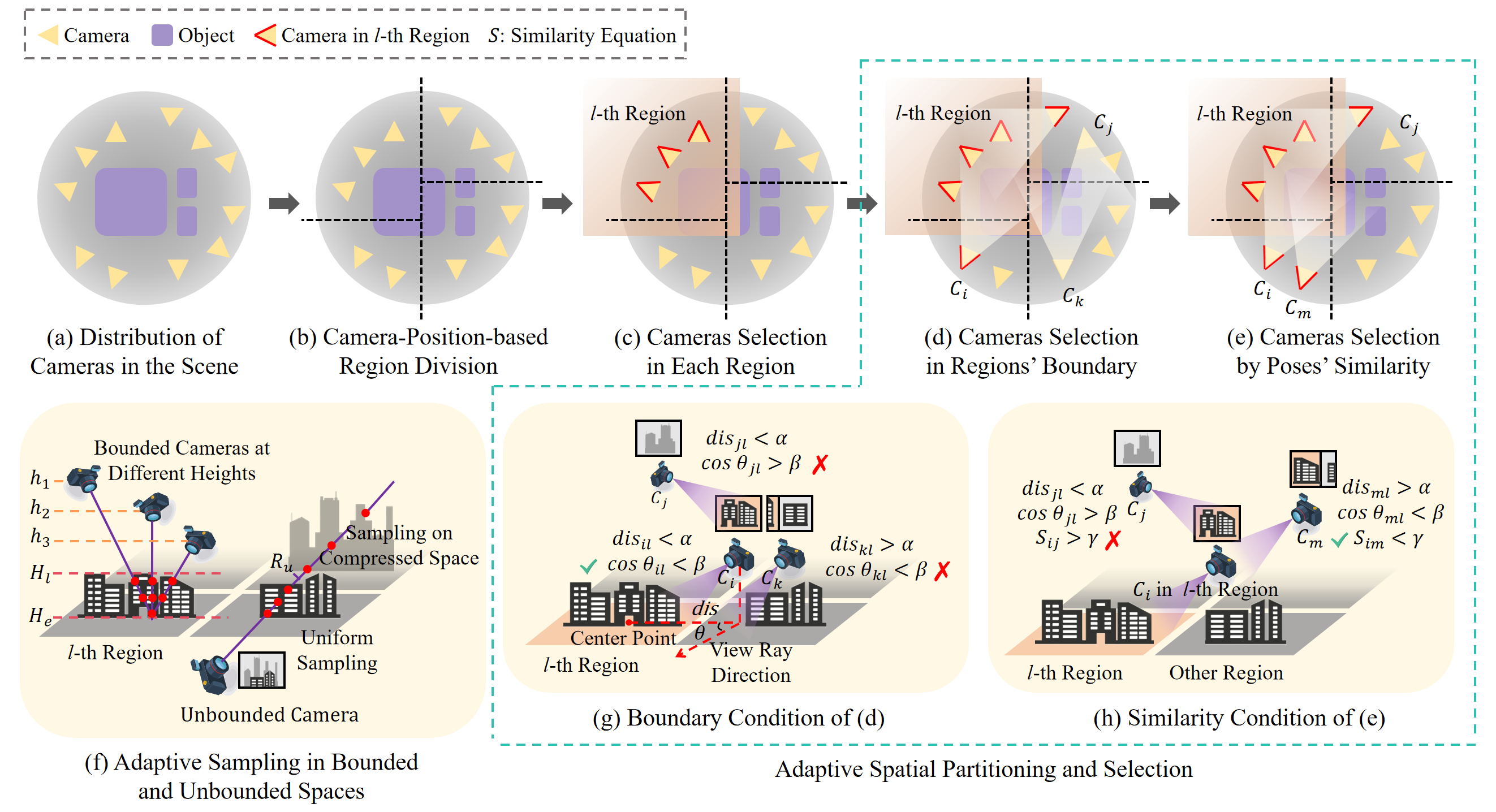}
\caption{
The pipeline of our method. We propose an adaptive spacial partitioning and selection method that makes our method applicable to aerial datasets of different 
trajectories.
(a) (b) divide the entire scene into multiple areas based on the poses of drones. 
Next, we select cameras that can observe $l$-th region, and use these cameras to train the NeRF of $l$-th region.
(c) selects the cameras in $l$-th region.
(d) selects the boundary cameras that can see the $l$-th region.
(e) utilizes the boundary cameras to select more cameras that can view the $l$-th region.
(f) samples between $H_1$ and $H_2$ in bounded space, and samples on buildings to infinity in unbounded space.
(g) is the condition for determining whether the boundary cameras belongs to the $l$-th region. 
When the distance $dis$ and angle $\theta$ are within the threshold, it indicates that this boundary camera can observe the $l$-th region.
(h) is to determine whether the non-boundary cameras outside the $l$-th region belongs to this region. 
We use the similarity equation to find cameras in other regions that are similar to the boundary camera of the $l$-th region, which indicates that 
these cameras can also observe $l$-th region.
}
\label{process}
\end{figure*}

The pipeline of our method is shown in Fig. 3. We partition the space into multiple regions based on the distribution of drones and select
corresponding cameras in each region based on boundary and similarity conditions.
We also propose an adaptive sampling algorithm that allows the sampling range to cover the buildings and samples near the buildings.

\subsection{Neural Radiance Field}

We use the original NeRF \cite{mildenhall2021nerf} as our network architecture. NeRF models a scene using a consistent 
volumetric radiance field to capture the scene's geometry and appearance variations. 
During rendering, NeRF calculates a view ray for each pixel based on the camera pose, and performs sampling on this view ray.
Then, the coordinates of sampling points and the direction of the view ray are passed through MLP to obtain the color $\mathbf{c}_i = (r, g, b)$ and density 
$\sigma_i$ of these 
sampling points. Finally, NeRF derives the pixel color $\hat{C}(\mathbf{r})$ through the integration:

\begin{equation}
\label{nerf_int}
\hat{C}(\mathbf{r})=\sum_{i = 0}^{N-1}T_i(1-exp(-\sigma_i\delta_i))\mathbf{c}_i
\end{equation}

\noindent where $\delta_i$ is the distance between samples $p_i$ and $p_{i+1}$. $T_i$ represents the cumulative transparency of $p_i$, and

\begin{equation}
\label{nerf_int_T}
T_i=exp(-\sum_{j = 0}^{i-1}\sigma_j\delta_j)
\end{equation}

NeRF employs a two-stage hierarchical sampling procedure to sample on view rays.
In a coarse stage, uniform sampling is performed within the sampling range, while in a fine stage, inverse distribution sampling is 
performed based on the density of the sampling points from the coarse stage.
During the training process, the model is minimized by the 
loss function $L_{mse}$ with the ground truth $C(\mathbf{r})$:

\begin{equation}
\label{nerf_int_loss}
L_{mse}=\sum_{\mathbf{r\in R}} \| C(\mathbf{r})-\hat{C}(\mathbf{r}) \| 
\end{equation}

\noindent where $R$ represents the batches of pixels. However, NeRF computes the color for each pixel independently, which leads to a loss of structural information in the images. 
We incorporate the $S3IM$ loss function \cite{xie2023s3im} $L_{S3IM}$ to constrain the structural information:

\begin{equation}
\label{S3IM_loss}
L_{S3IM}=1-\frac{1}{M}\sum_{m = 1}^{M}SSIM(\mathbb{P}^{(m)}(\hat{C}(\mathbf{r})),\mathbb{P}^{(m)}(C(\mathbf{r})))  
\end{equation}

The final loss function $L$ is defined as:

\begin{equation}
\label{loss}
L=\lambda_{MSE}L_{MSE}+\lambda_{S3IM}L_{S3IM}
\end{equation}

\noindent where $\lambda_{MSE}$, $\lambda_{S3IM}$ are the weights of $L_{MSE}$, $L_{S3IM}$.

\subsection{Spatial Partitioning and Selection}

\begin{figure*}[!t]
  \centering
  \includegraphics[width=6in]{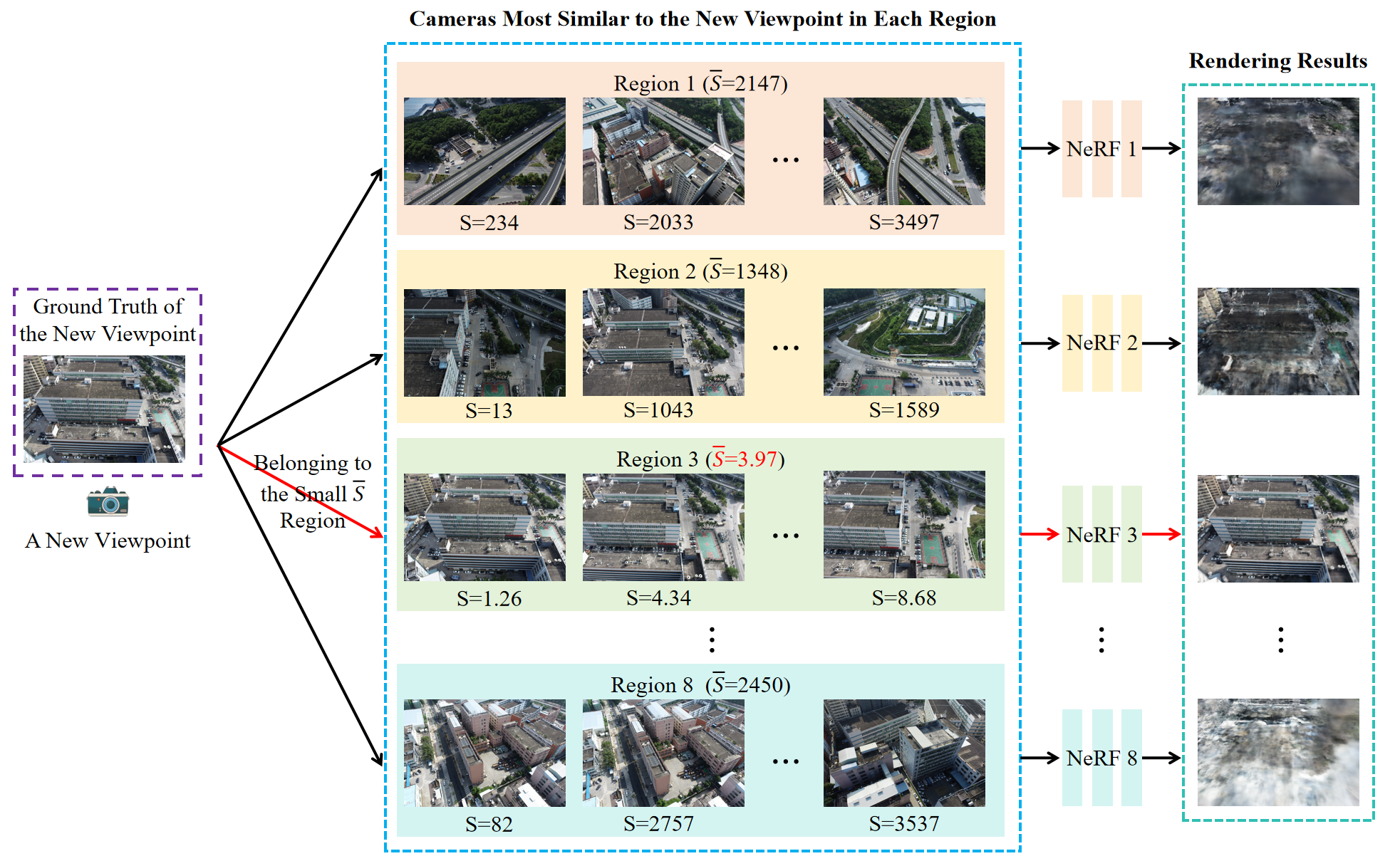}
  \caption{Visualization of our spatial selection strategy. The algorithm's input is the pose of a new viewpoint, and the output is the rendering image of this viewpoint. 
  In each region, we find the $n_s=5$ cameras with the smallest $S$ (calculated by (7)) relative to this viewpoint. The smaller the $S$, the higher the similarity between cameras, 
  and the more common view areas there are. For instance, the scene viewed from the new viewpoint is almost identical to that captured by a camera with $S=1.26$.
  To avoid randomness, taking the average of $n_s=5$ cameras' $S$ as the region similarity error $\overline{S}$ between this 
  viewpoint and each region. When $\overline{S}$ is small, this camera belongs to this region, and the NeRF of this region is used to render this viewpoint.
  As can be seen from the images, the smaller the $\overline{S}$, the smaller the difference between images of the new viewpoint and this region's cameras, indicating that the new viewpoint belongs to this space.
  }
  \label{sapce_sel}
  \end{figure*}
  
We divide the scene into multiple regions based on the distribution of drones, and then
perform clustering on the XY-plane using k-means clustering method \cite{ahmed2020k}.
The scene is divided into N regions $cluster_i, i=1,2...,N$, and $N$ region centroid points $\mathbf{o}_i, i=1,2...,N$ ($\mathbf{o}_i$ is the coordinate 
on the XY-plane) are obtained. There may be cameras from $cluster_j$ that can see $cluster_i$.
We aim to train the $i$-th region's NeRF using enough
cameras which can observe a certain region. 
Therefore, we need to observe the boundary cameras between regions. The characteristic of boundary cameras is that they are close to other regions 
and may be able to view other regions' scenes. For $cluster_j$, if there exists a camera whose coordinate is denoted as $\mathbf{p}_j$ that satisfies $\|\mathbf{p}_j-\mathbf{p}_i\|<\alpha$ with a 
camera $\mathbf{p}_i$ in $cluster_i$ and a threshold $\alpha$, then the camera $\mathbf{p}_j$ is considered as a boundary camera between $cluster_i$ and $cluster_j$.

Next, we select the boundary cameras that can observe $cluster_i$.
If the boundary camera $\mathbf{p}_j$ is in $cluster_j$, it needs to be determined whether this camera can view $cluster_i$. The orientation 
of this camera in its coordinate system is $\mathbf{d}_c=(0,0,1)$. We denote the rotation and translation from the camera coordinate system to the world coordinate 
system as $\mathbf{R}$ and $\mathbf{t}$, respectively. Then, the orientation of the camera in the world coordinate system $\mathbf{d}_w$ is:

\begin{equation}
\label{ori_c2w}
\mathbf{d}_w=\mathbf{R}\mathbf{d}_c+\mathbf{t}
\end{equation}

We denote the vector pointing from camera $\mathbf{p}_j$ to $\mathbf{o}_i$ as $\mathbf{a}$. If the angle $\theta$ between $\mathbf{d}_w$ and $\mathbf{a}$ 
satisfies $cos\theta>0$, it indicates that camera $\mathbf{p}_j$ can observe region $cluster_i$ and then add this camera to $cluster_i$. 
Since this camera can also view the scenes from $cluster_j$, it belongs to both $cluster_i$ and $cluster_j$.

Due to the limited number of boundary cameras, there exist some cameras in $cluster_j$ that, although are not boundary cameras, can view $cluster_i$.
These cameras need to undergo region redefinition as well. For two cameras $\mathbf{p}_1$ and $\mathbf{p}_2$,
we use the rotation $\mathbf{R}$, translation $\mathbf{t}$, 
and capture time $time$ to measure their similarity. $\mathbf{R}$ and $\mathbf{t}$ help to keep the similarity of the cameras' spatial position and orientation.
$time$ ensures that the shooting times are not too far apart and maintain small differences in light changes between cameras.
The similarity error $S$ between camera $\mathbf{p}_1$ and camera $\mathbf{p}_2$ is calculated as:

\begin{equation}
\label{similarity}
S= \left\lVert \begin{bmatrix} \mathbf{R}_1 & \mathbf{t}_1 \\ 0 & time_1 \end{bmatrix} - \begin{bmatrix} \mathbf{R}_2 & \mathbf{t}_2 \\ 0 & time_2 \end{bmatrix} \right\rVert 
\end{equation}

\begin{figure}[!t]
  \centering
  \includegraphics[width=2.5in]{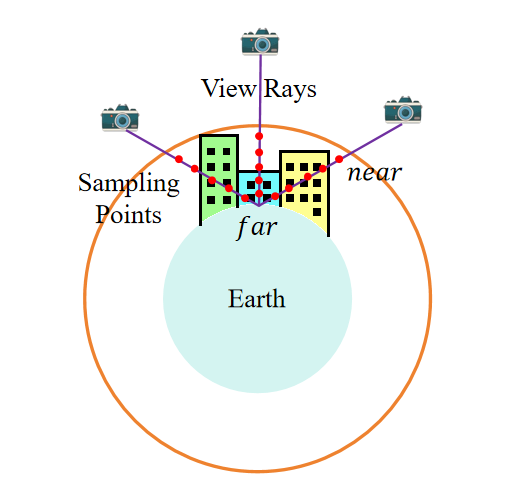}
  \caption{Sampling strategy on bounded regions. The intersection of the drone's ray with the outer sphere $near$ is the starting point for sampling, and the 
  intersection with the Earth $far$ is the ending point for sampling. Each drone's ray is sampled to cover buildings on the ground.}
  \label{sam_boud}
  \end{figure}

\begin{figure}[!t]
  \centering
  \includegraphics[width=3in]{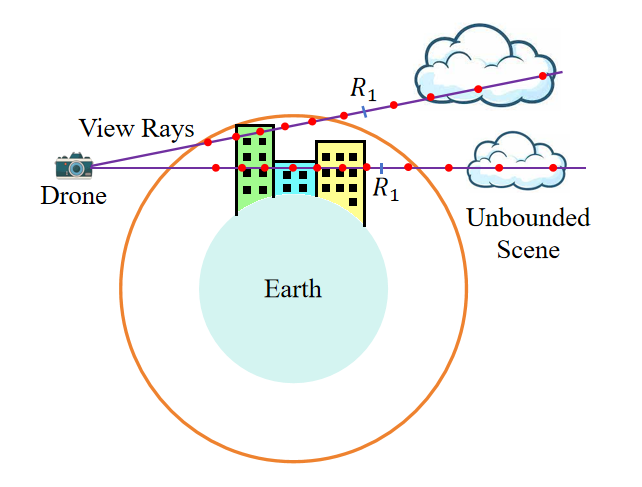}
  \caption{Sampling strategy for unbounded regions. When the view rays do not intersect with the Earth or are directed towards the sky, objects 
  in unbounded regions, such as clouds, are visible. We uniformly sample from the camera origin to $R_1$. The sampling interval gradually increases in the
  $[R_1,+\infty]$.}
  \label{sam_unboud}
  \end{figure}

The smaller the value of $S$, the more similar the two cameras are. For each boundary camera, we calculate $n_p$ cameras with the most minor similarity error.
Then, they are assigned to the same region as the corresponding boundary cameras.
On the basis of that, we divide the entire large-scale scene into multiple regions and train a 
separate NeRF on each region. The advantage of dividing space based on drones' pose is that our model can be applied to any aerial photography trajectory and
the computational resources do not increase with the scene's size.
  
\textbf{Spatial Selection for New Viewpoints.} During training, we divide the cameras into corresponding regions and train the NeRF for each region. During inference, 
For a new viewpoint, we creatively design an accurate and fast spatial selection method based on the known cameras in the space to determine which 
region it belongs to and then render this viewpoint with the corresponding NeRF, as shown in Fig. 4.
Firstly, we should determine $n_s$ cameras most similar to the new viewpoint at each region by computing the similarity error $S$ in equation (7),
and take the mean of $n_s$ cameras as the region similarity error $\overline{S}$ between the new viewpoint and each region. 
If the $\overline{S}$ of a particular region satisfies $\overline{S}<\gamma$, where $\gamma$ is the threshold, 
It indicates that the new viewpoint can observe the scene of this region.
Therefore, the new viewpoint is rendered by this region's NeRF.
When the above condition is satisfied for several regions, 
we simply render the new viewpoint with the NeRFs of these regions and then take the average of all rendering results to get the final rendering result.

\subsection{Adaptive Sampling}

During the process of training a NeRF for a region, sampling is performed along each ray for every pixels. The previous sampling methods do not cover the object adequately, 
resulting in wasted sampling points and loss of accuracy. We propose an adaptive sampling method to ensure cameras at different heights sampling near objects. 

\textbf{Sampling on Bounded Space.} As shown in Fig. 5, we assume the Earth is a sphere with a radius of $R_{Earth}$. The height $h$ of the tallest building 
in the current region can be obtained from the sparse point cloud calculated by COLMAP \cite{fisher2021colmap}. A new sphere with the center at the center of 
the Earth and a radius of $R = R_{Earth} + h$ is constructed. A camera's position in space is given by $(x_0, y_0, z_0)$, and its view ray direction is represented 
by $(a, b, c)$. The parameter equation of this view ray with respect to t can be expressed as:

\begin{equation}
\begin{cases}
x=x_0+at \\ 
y=y_0+bt \\
z=z_0+ct \\
\end{cases}
\end{equation}

The center coordinate of the Earth is $(r_1, r_2, r_3)$, then the Earth can be built as:

\begin{equation}
(x-r_1)^2+(y-r_2)^2+(z-r_3)^2=R_{Earth}^2
\end{equation}

The outer sphere can be represented as:

\begin{equation}
(x-r_1)^2+(y-r_2)^2+(z-r_3)^2=R^2
\end{equation}

The intersection of the view ray with the outer sphere is the nearest sampling point "$near$", and the intersection with the Earth is 
the farthest sampling point "$far$". Substituting equation (8) into equations (9) and (10), $near$ and $far$ can be solved by

\begin{equation}
near=\frac{-B-\sqrt{B^2-4AC_{Earth}} }{2A}
\end{equation}

\begin{equation}
far=\frac{-B-\sqrt{B^2-4AC} }{2A}
\end{equation}

\noindent where 

\begin{gather*}
A=a^2+b^2+c^2 \\ 
B=2(a(x_0-r_1)+b(y_0-r_2)+c(z_0-r_3)) \\
C_{Earth}=(x_0-r_1)^2+(y_0-r_2)^2+(z_0-r_3)^2-R_{Earth}^2 \\
C=(x_0-r_1)^2+(y_0-r_2)^2+(z_0-r_3)^2-R^2 
\end{gather*}

Thus, the sampling range along the view ray is $[near, far]$. Similar to NeRF \cite{mildenhall2021nerf}, we first uniformly sample within this range, then calculate the 
density of the sampled points, and perform inverse distribution sampling based on the density.

\textbf{Sampling on Unbounded Space.} When the camera observes unbounded regions such as the sky and clouds, the view ray does not intersect with the 
Earth. Therefore, we design a sampling method for unbounded regions, as shown in Fig. 6. We divide the view ray into 
foreground $[0, R_1]$ and background $[R_1,+\infty]$. We uniformly sample in the foreground, and sample points gradually further away on the 
background, extending to unbounded space. To that end, we compress the space as:

\begin{equation}
cam(t) =\begin{cases}
t, 0 \le  t \leq  R_1\\ 
R_1+\frac{1}{R_1}-\frac{1}{t}, otherwise \\
\end{cases}
\end{equation}
  
\noindent where $cam(t)$ represents the compressed space. When $t > R_1$, we sample uniformly in the compressed space. Let the sampling point be s, $s\in (R_1,R_1+\frac{1}{R_1})$, then 
the corresponding sampling point mapped back to the original space is:

\begin{equation}
t=\frac{1}{-s+R_1+\frac{1}{R_1}}
\end{equation}

Hence, the sampling range for t is $(R_1, +\infty)$, and the space between sampling points becomes increasingly larger.

\section{Experiments}

\begin{figure}[!t]
  \centering
  \includegraphics[width=3.3in]{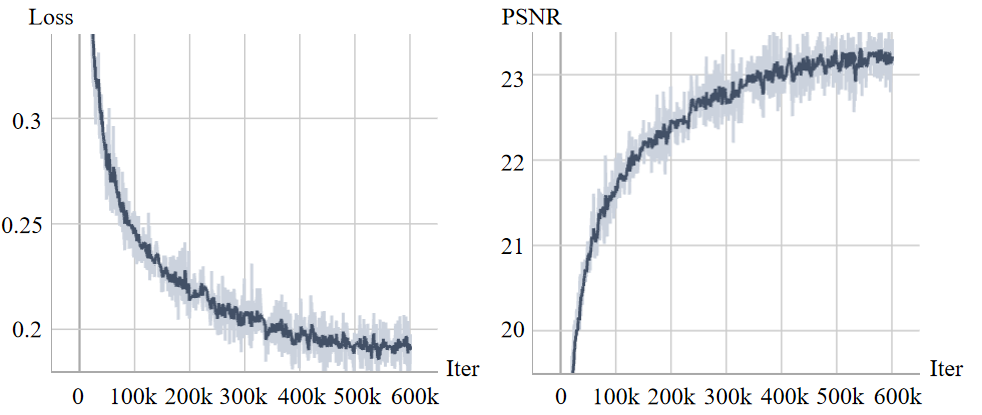}
  \caption{The relationship between the number of iterations, loss function, and PSNR. The model converges after 600k iterations.}
  \label{loss_pic}
  \end{figure}

\subsection{Datasets}

\textbf{56 Leonard} \cite{xiangli2022bungeenerf} consists of aerial datasets taken at different altitudes. When collecting data, the strategy is to move
the camera in a circular motion and gradually elevate the camera from a low altitude to a high altitude. The range of cameras in this dataset is $300m \times 300m$.

\textbf{Residence} from UrbanScene3D \cite{liu2021urbanscene3d} consists 
of large-scale scene pictures captured at the same altitude with a uniform camera trajectory. 
The range of cameras in this dataset is $250m \times 400m$.

\textbf{SCUTic} is captured at the same altitude utilizing uneven aerial trajectories.
We conduct aerial photography around each building. In areas where buildings are dense, the density of cameras is high, while in areas where 
buildings are sparse, the density of cameras is low.
This dataset contains 5.86 GB high-resolution images taken from the South China University of Technology International Campus by a DJI Mini 2 drone with $500m \times 600m$ 
camera range. 
The uneven cameras' distribution and the large camera range pose challenges for view rendering and 3D reconstruction to test the robustness and generalizability 
of different methods.

\begin{figure*}[!t]
  \centering
  \includegraphics[width=7in]{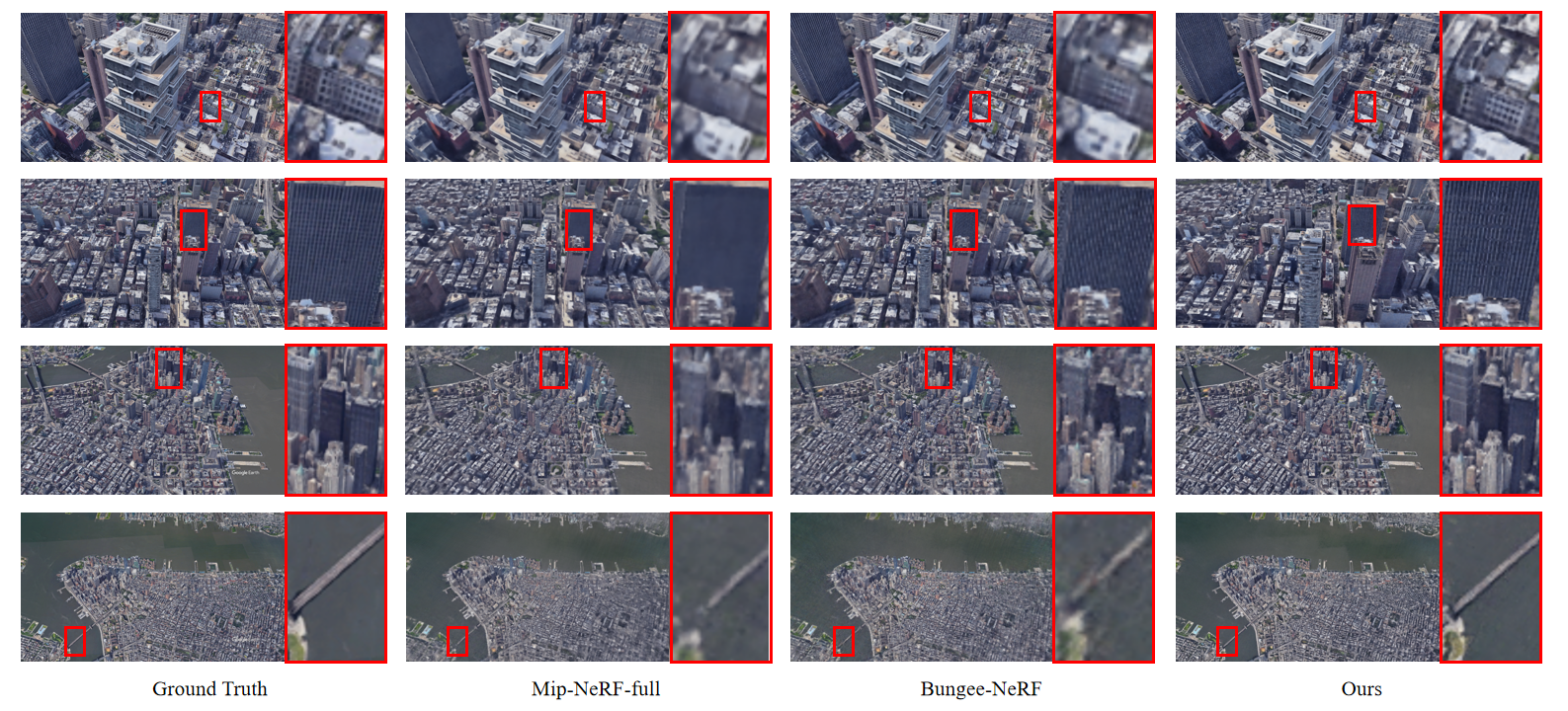}
  \caption{The comparison of rendering results on four different altitudes. Our rendering results show clearer texture details at 
  lower altitude and more complete image information at higher altitude.}
  \label{56_com}
  \end{figure*}

\begin{figure}[!t]
\centering
\includegraphics[width=3.3in]{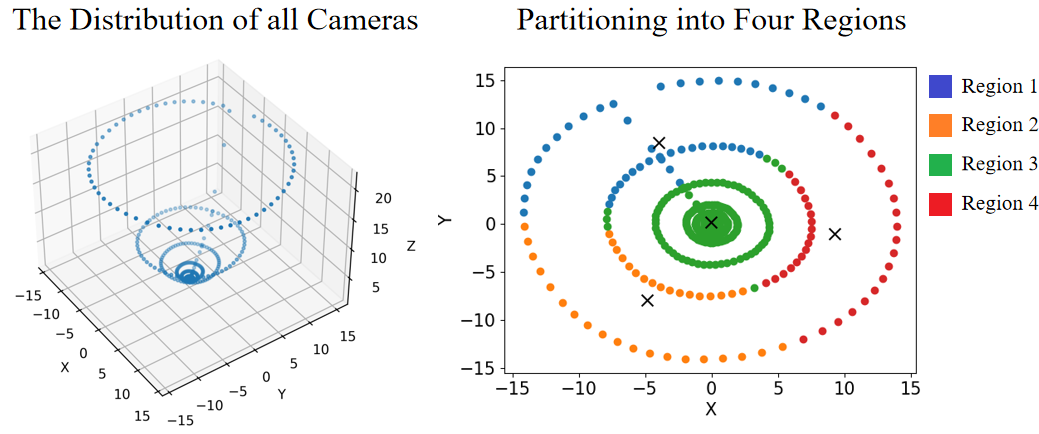}
\caption{Our spatial partitioning on the 56 Leonard dataset. 
The left figure represents the distribution of drones in this dataset and the right figure shows the cameras we selected in each region.
X, Y and Z axes represent the range of drone distribution, with a unit of 10 meters.
Those cameras at low altitudes are relatively close and are grouped into one region. 
At high altitudes, cameras at different heights are grouped into one region.}
\label{56_par}
\end{figure}

\subsection{Metrics and Settings}

\textbf{Metrics.} Our results on novel view synthesis are quantitatively evaluated by PSNR \cite{hore2010image}, SSIM \cite{wang2004image}, and the 
LPIPS implementation of VGG \cite{zhang2018unreasonable}. PSNR is utilized to calculate the mean squared error between two images in 
logarithmic space. SSIM is more concerned with structural similarity. LPIPS is used to assess perceptual similarity.

\textbf{Settings.} The scene is divided into several regions and a separate NeRF is trained on each region. For the Residence dataset, the scene is partitioned
into 8 regions. For the 56 Leonard and SCUTic datasets, the scene is divided into 4 regions. Similar to NeRF \cite{mildenhall2021nerf}, the 8-layer MLP is utilized for feature extraction, where each layer 
produces features with 256 channels. The 48-dimensional appearance encoding is employed to adapt the model to different lighting conditions. We train 600k iterations 
for each dataset with 4096 batch size. The loss and PSNR versus the number of iterations are shown in Fig. 7, respectively. The Adam is applied as optimizer \cite{kingma2014adam} with a learning rate decaying exponentially from $5\times10^{-4}$
to $5\times10^{-5}$. The number of sampling points at the coarse sampling stage is 64, and at the fine stage is 128. Our model is trained by a single RTX 3090.

\subsection{Results}

\begin{table}[!t]
  \caption{Comparison of Different Methods on the 56 Leonard Dataset. The Bold Data in each Column Represents the Best for each Metric.\label{tab:56_com}}
  \centering
  \begin{tabular}{|c c c c|}
  \hline
  Method & PSNR$\uparrow$ & SSIM$\uparrow$ & LPIPS$\downarrow$ \\
  \hline
  NeRF \cite{mildenhall2021nerf} (D=8,Skip=4) & 21.702 & 0.320 & 0.636 \\
  Mip-NeRF-small \cite{barron2021mip} & 23.337 & 0.709 & 0.354 \\
  Mip-NeRF-large \cite{barron2021mip} & 23.507 & 0.718 & 0.346 \\
  Mip-NeRF-full \cite{barron2021mip} & 23.665 & 0.732 & 0.328 \\
  Bungee-NeRF \cite{xiangli2022bungeenerf} & 24.513 & 0.815 & 0.160 \\
  \hline
  Ours & \textbf{25.333} & \textbf{0.832} & \textbf{0.148} \\
  \hline
  \end{tabular}
  \end{table}

  \begin{figure}[!t]
    \centering
    \includegraphics[width=3.3in]{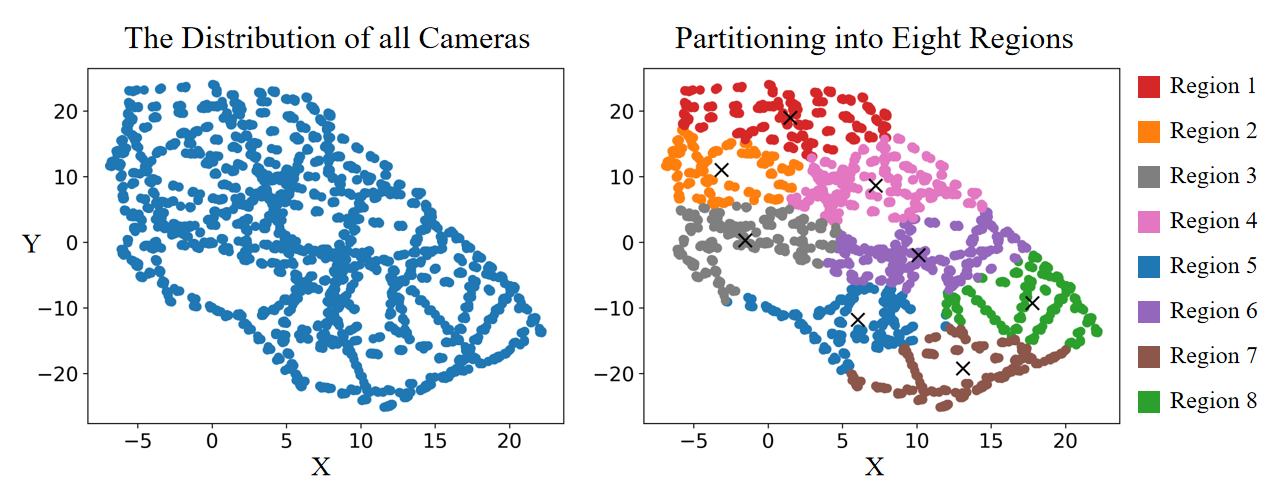}
    \caption{Our spatial partitioning on the Residence dataset.
    The left figure represents the distribution of drones on the 
    XY-plane, while the right figure illustrates the clustering of cameras.
    The X and Y axes represent the range of drone distribution, with a unit of 10 meters.
    We divide the cameras relatively evenly into the corresponding space.
    }
    \label{k-means}
    \end{figure}

\begin{figure*}[!t]
  \centering
  \includegraphics[width=7in]{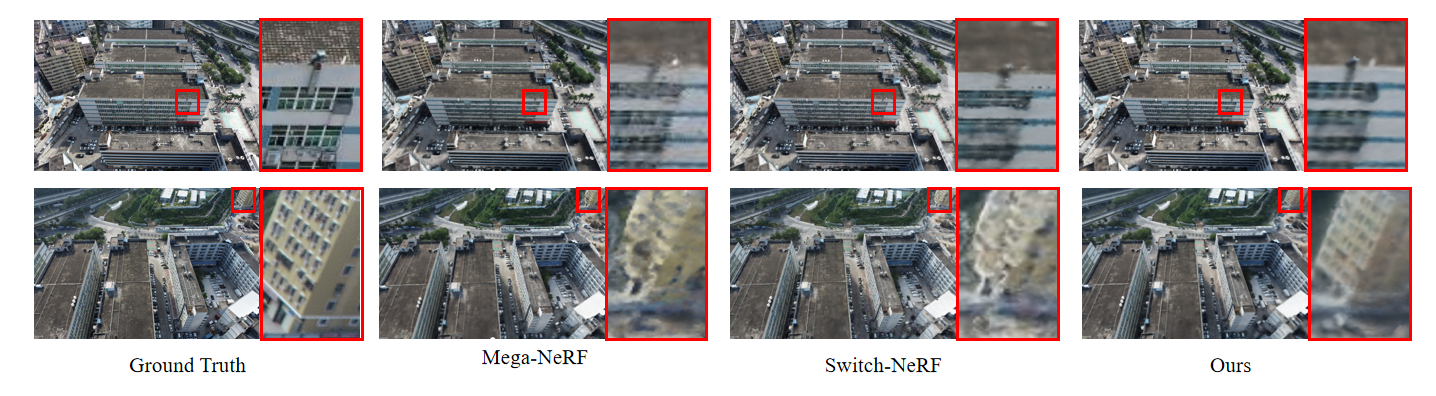}
  \caption{Comparison of results on the Residence dataset. Our designed adaptive sampling and spatial partitioning strategy can improve rendering accuracy.}
  \label{res_results}
  \end{figure*}

\begin{table*}[!t]
  \caption{Comparison of Different Methods on the Residence Dataset.
  Extensibility Indicates Whether a Single GPU Can Render Scenes of any Size.
  The Bold Data in each Column Represents the Best for each Metric.\label{tab:res_com}}
  \centering
  \begin{tabular}{c c c c c c c}
  \hline
  Method & PSNR$\uparrow$ & SSIM$\uparrow$ & LPIPS$\downarrow$ & GPU Memory$\downarrow$ & Rendering Time$\downarrow$ & Extensibility\\
  \hline
  NeRF \cite{mildenhall2021nerf} & 19.01 & 0.593 & 0.488  & 26G & \textbf{41s} & $\times$  \\
  NeRF++ \cite{zhang2020nerf++} & 18.99 & 0.586 & 0.493  & 29G & 44s & $\times$ \\
  Mega-NeRF \cite{turki2022mega} & 22.08 & 0.628 & 0.489  & 10G & 246s & \checkmark \\
  Switch-NeRF \cite{zhenxing2022switch} & 22.57 & 0.654 & 0.457 &  32G & 200s & $\times$\\
  \hline
  Ours & \textbf{23.52} & \textbf{0.742} & \textbf{0.261} & \textbf{8G} & 48s & \checkmark\\
  \hline
  \end{tabular}
  \end{table*}

\textbf{Results on 56 Leonard.} The comparison 
of rendering results at four different altitudes is shown in Fig. 8. 
Our method achieves rendering the texture of buildings more clearly in low-altitude areas and structural information more completely in high-altitude areas, respectively.
Bungee-NeRF \cite{xiangli2022bungeenerf} classifies cameras at each altitude into one category, and uses a coarse-to-fine method to train NeRFs from high-altitude areas to 
low-altitude areas sequentially. However, this space partitioning method based on altitude can result in a significant loss of texture information.
The reason is that the texture details decrease as the altitude increases. Bungee-NeRF \cite{xiangli2022bungeenerf} only utilizes images from one altitude to 
train the corresponding NeRFs, resulting in less information to describe each building, and blurry and incomplete rendering results.
Our adaptive spatial method effectively solves this problem by grouping cameras at different altitudes into one category, as shown in Fig. 9.
At the lowest altitude, the cameras are densely distributed, and there is a lot of shared information between adjacent cameras. Grouping them into 
one category can yield accurate rendering results.
As the altitude increases, the range of the scene expands while the texture details of the buildings decrease. In this case, grouping cameras at 
different heights into one category provides both global structure information and texture details, thereby improving the quality of rendering.

From Table \Rmnum{1}, as compared to
NeRF \cite{mildenhall2021nerf} and Mip-NeRF \cite{barron2021mip}, our method achieves 1.668, 0.1, and 0.18 dB increase in PSNR, SSIM, and LPIPS, respectively.
Compared to Bunggee-NeRF \cite{xiangli2022bungeenerf} with 120 hours for training, ours achieves 0.82, 0.017, and 0.012 dB increase in PSNR, SSIM, and LPIPS, 
respectively, and only requires 32 tranining hours which indicates the effectiveness of our adaptive spatial partitioning algorithm.

\textbf{Results on Residence.} 
Fig. 10 shows the results of our spatial partitioning on the Residence dataset, where
the pose distribution of drones is relatively uniform. In this case, various models can achieve high-quality rendering.
Fig. 11 compares the rendering results of our method with other approaches.
The previous methods do not render the texture details finely enough and result in incomplete rendering of distant buildings.
The reason is that these methods set the sampling range as a hyperparameter and the sampling points cannot cover the entire scene, 
affecting the quality of rendering.
We design the adaptive sampling method which ensures that the sampling range at different altitudes
covers buildings to deal with this issue.  

In Table \Rmnum{2}, as
compared to the SOTA Switch-NeRF \cite{zhenxing2022switch}, our method uses 1/4 sampling points to save 24 GB GPU memory but 
improves 0.95, 0.088, and 0.196 dB in PSNR, SSIM, and LPIPS, respectively.
In addition, we compare the rendering time of different methods. 
Since NeRF \cite{mildenhall2021nerf} and NeRF++ \cite{zhang2020nerf++} do not perform spatial partitioning, there is no need to select regions for new perspectives, 
and they achieve fast rendering.
Mega-NeRF \cite{turki2022mega} applies the NeRFs on 
regions where a ray passes through to jointly calculate the color of a pixel. This results in a linear increase in rendering time as the number of regions 
increase. Switch-NeRF \cite{zhenxing2022switch} requires an expert network to determine which region each sampling point belongs to, and
then calculates the color and density 
of the sampling point with the corresponding NeRF. The large expert network slows down the rendering speed.
Our method creatively utilizes existing cameras' poses to determine which region a new viewpoint belongs to, 
which does not rely on a network and dramatically speeds up the rendering.
Compared with Mega-NeRF \cite{turki2022mega} and Switch-NeRF \cite{zhenxing2022switch}, our method increases the rendering 
speed by 4 times.
We also compare the extensibility of different methods if they can enable to render arbitrarily large scenes using a single GPU.
NeRF \cite{mildenhall2021nerf}, NeRF++ \cite{zhang2020nerf++} and Switch-NeRF \cite{zhenxing2022switch} train all images at once and the GPU memory increases as the scene size increases.
Mega-NeRF \cite{turki2022mega} first divides the whole scene into several regions and trains the corresponding NeRFs using each region's images, which significantly 
saves the GPU memory. That is to say, as long as the spatial division method is appropriate, any large aerial scene can be rendered based on 
a single GPU. 
However, Mega-NeRF \cite{turki2022mega} divides the space evenly into multiple parts. The rendering accuracy is significantly lower down when the distribution of the 
drones' trajectory is uneven. Our spatial partitioning and selection method based on drone pose distribution answers in the affirmative to address that, and is 
robust to the cameras' trajectory of aerial datasets.

\begin{figure*}[!t]
  \centering
  \includegraphics[width=7in]{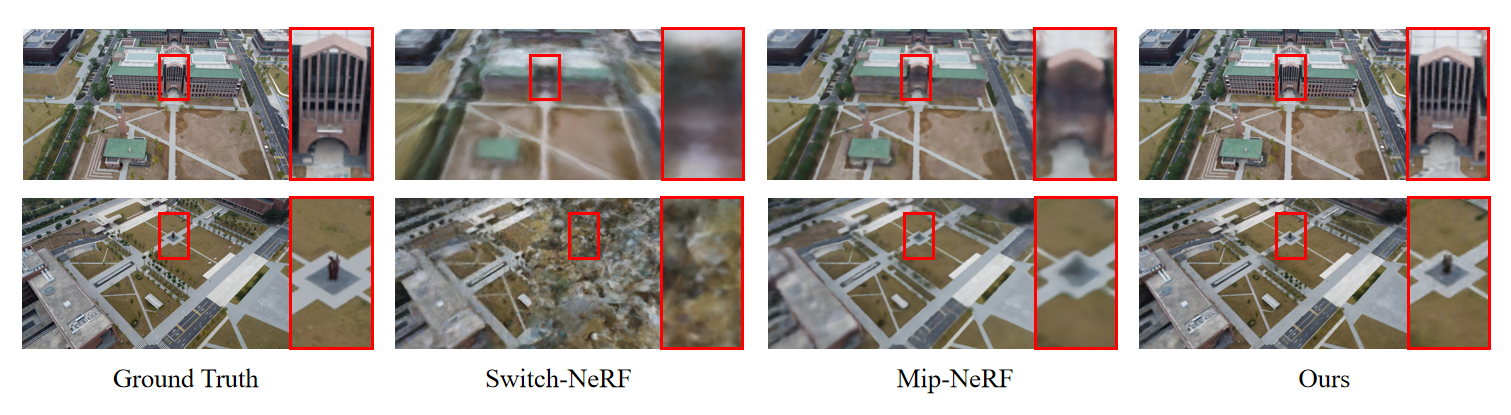}
  \caption{Comparison of results on the SCUTic dataset. When the distribution density of drones in space is uneven, our spatial partitioning method based on camera 
  poses is more robust compared to previous spatial partitioning methods.}
  \label{SCUT_res}
  \end{figure*}

\begin{figure}[!t]
  \centering
  \includegraphics[width=3.4in]{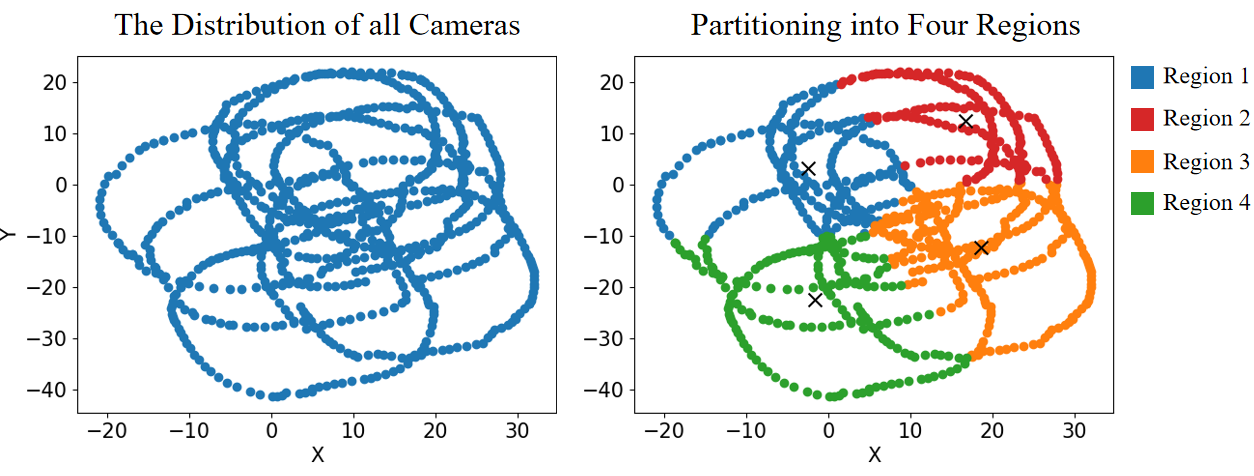}
  \caption{The distribution of drones and region division of our method on the SCUTic dataset. This dataset is obtained by drones circling around each 
  building. When the buildings are dense, the camera distribution is dense. When the buildings are sparse, the camera distribution is sparse.
  We evenly divide the cameras into their respective regions based on our spatial partitioning method.}
  \label{SCUT_par}
  \end{figure}

\textbf{Results on SCUTic.} 
Existing aerial datasets have a relatively uniform 
distribution of cameras in space, and various methods can achieve satisfactory rendering results.
To verify the robustness of those methods, we create a new dataset, named SCUTic, in a way that the camera distribution is uneven. 
The rendering results on this dataset are shown in Fig. 12. 
It can be seen that the SOTA method Switch-NeRF \cite{zhenxing2022switch} is easily influenced by the aerial flight trajectory.
The reason is that the uneven aerial flight trajectory results in a large difference in information across different regions. 
Switch-NeRF \cite{zhenxing2022switch} needs to integrate the uneven information from those regions when view rays pass through to synthesize a new viewpoint.
We train a separate NeRF for each area to learn different spatial information.
Our spacial selection algorithm based on pose similarity can adaptively assign the new viewpoint to the corresponding region and use the NeRF of this region 
for rendering, thereby solving the problem of information difference.
The spatial partitioning of our method is shown in Fig. 13.
For example, in Region 2, the cameras are densely distributed and the amount of information is large, therefore the number of cameras assigned to this region 
is relatively fewer. In contrast, in Region 4, the cameras are sparse and the amount of information is small, therefore more cameras are allocated to increase 
the information in this region. The rendering performance of each area is improved by reasonably allocating cameras.

\begin{table}[!t]
  \caption{Comparison Results on the SCUTic Dataset. Partitioning Represents Spacial Partitioning and Selection Methods. The Bold Data in each Column Represents the Best for each Metric.\label{tab:scut}}
  \centering
  \begin{tabular}{|c c c c c|}
  \hline
  Method & Partitioning & PSNR$\uparrow$ & SSIM$\uparrow$ & LPIPS$\downarrow$ \\
  \hline
  Mip-NeRF \cite{barron2021mip}  & - & 22.40 & 0.717 & 0.318 \\
  Mega-NeRF \cite{turki2022mega} & Uniform & - & - & - \\
  Switch-NeRF \cite{zhenxing2022switch} & Learning & 18.8  & 0.479 & 0.598 \\
  \hline
  Ours & Pose &\textbf{27.62} & \textbf{0.847} & \textbf{0.108} \\
  \hline
  \end{tabular}
  \end{table}

In Table \Rmnum{3},
Mega-NeRF \cite{turki2022mega} partitions the space evenly, while Switch-NeRF \cite{zhenxing2022switch} divides the space through learning.
These two approaches are greatly influenced by the flight trajectory and Mega-NeRF \cite{turki2022mega} even fails to render.
Our method partitions the space based on the poses of drones to satisfy different flight trajectories.
Compared to Switch-NeRF \cite{zhenxing2022switch}, our method achieves 5.22, 0.368, and 0.49 dB increase in PSNR, SSIM, and LPIPS, respectively.
The rendering results of our method on all datasets are shown in Fig. 14. 

\section{Ablation Studies}

\begin{figure*}[!t]
  \centering
  \includegraphics[width=6.3in]{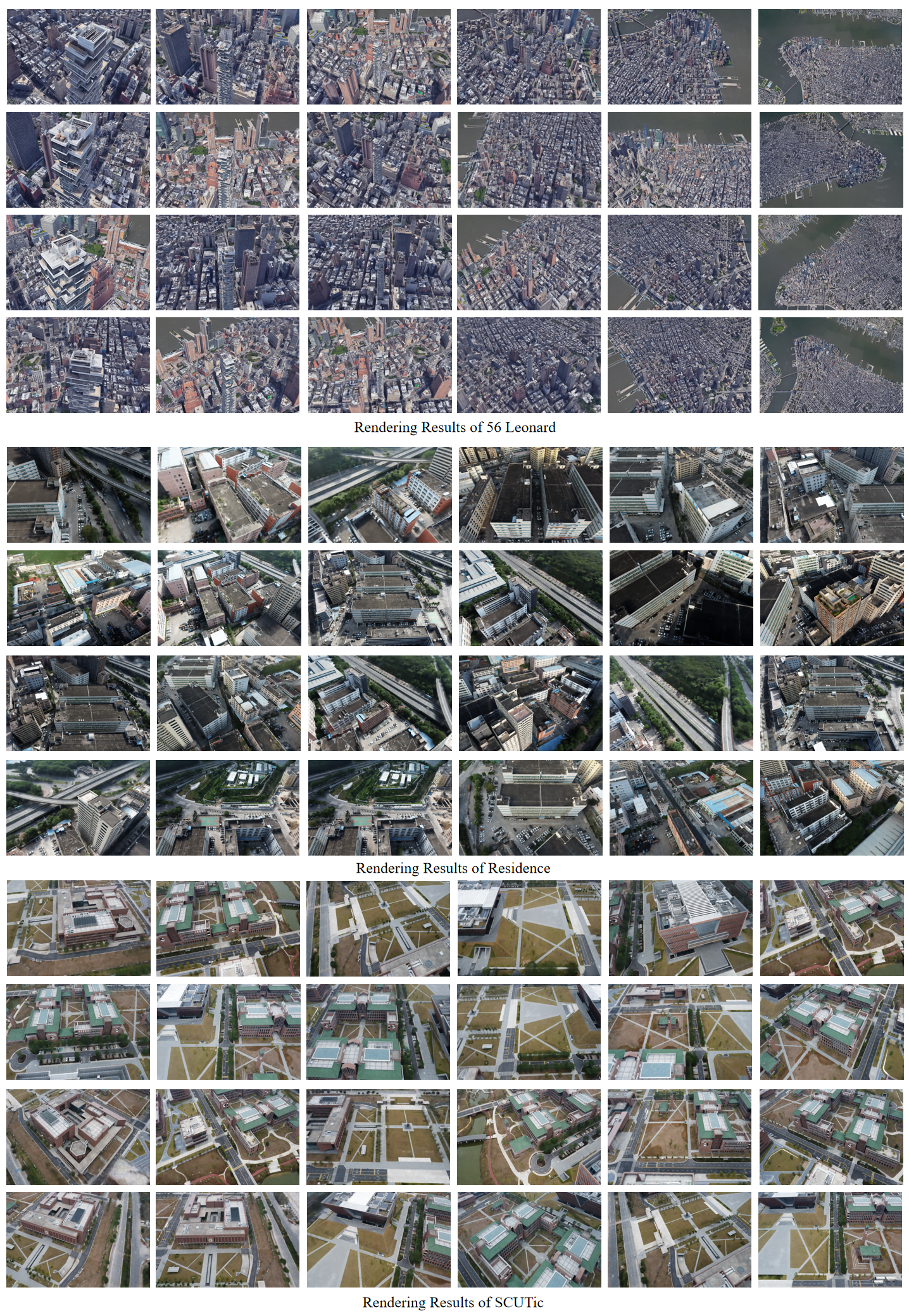}
  \caption{Rendering results of our method on three datasets. Our method can achieve vivid rendering results at the same height, different heights, and in 
  situations where the drone distribution is uneven.}
  \label{render_all}
  \end{figure*}

Our method partitions the scene based on the distribution of drones and trains 
a separate NeRF on each area. As shown in 
Table \Rmnum{4}, the more the number of divided regions is, the more texture details of each region's NeRF can perceive, and the higher the rendering accuracy is achieved. 

\begin{table}[!t]
\caption{Comparison of Rendering Results with Different Numbers of Region Partitions. The Bold Text in each Column Represents the Best Data for 
each Metric.\label{tab:abl_num}}
\centering
\begin{tabular}{|c c c c|}
\hline
Method & PSNR$\uparrow$ & SSIM$\uparrow$ & LPIPS$\downarrow$ \\
\hline
Ours-4 & 22.62 & 0.722 & 0.293 \\
Ours-8 & 23.52 & 0.742 & 0.261 \\
Ours-16 & \textbf{23.66} & \textbf{0.763} & \textbf{0.229} \\
\hline
\end{tabular}
\end{table}

\begin{table}[!t]
  \caption{Ablation Experiments for each Module of Our Method. The Bold Data in each Column Represents the Best for 
  each Metric.\label{tab:abl_mou}}
  \centering
  \begin{tabular}{|c c c c|}
  \hline
  Method & PSNR$\uparrow$ & SSIM$\uparrow$ & LPIPS$\downarrow$ \\
  \hline
  w/o Spacial Partitioning & 22.21 & 0.718 & 0.326 \\
  w/o Spacial Selection & 22.97 & 0.728 & 0.287 \\
  w/o Adaptive Sampling & 21.36 & 0.647 & 0.357 \\
  \hline
  Full Method & \textbf{23.52} & \textbf{0.742} & \textbf{0.261} \\
  \hline
  \end{tabular}
  \end{table}

Table \Rmnum{5} shows the effect of each module in the proposed pipeline. They are:

\textbf{Spacial Partitioning.} Our method divides the space based on the pose of cameras, which can reasonably distribute the 
information of the scene to each region and improve the rendering effect. 
Compared with evenly dividing the scene into several areas, our method achieves 1.31, 0.024, 0.065 dB increase in PSNR, SSIM and LPIPS, respectively.

\textbf{Spacial Selection.} We design the boundary condition and similarity condition to select cameras for a certain region. 
Compared with only considering camera coordinates to allocate cameras, our method achieves 0.55, 0.014, 0.026 dB increase in PSNR, SSIM and LPIPS, respectively.

\textbf{Adaptive Sampling.} Our method can make the sampling range cover the entire buildings and distribute numerous sampling points on the buildings, further
improving the rendering quality. Compared with setting the sampling range as a hyperparameter, our method achieves 2.16, 0.095, 0.096 dB increase in PSNR, 
SSIM and LPIPS, respectively.

\section{Conclusion}

In this paper, we propose an efficient and robust rendering method, termed Aerial-NeRF, for processing large-scale aerial datasets. 
Herein, our method using an adaptive spatial partitioning and selection approach
outperforms existing large-scale aerial rendering competitors by a large margin on the rendering speed, almost at 4 times.
Additionally, under the appropriate number of divided regions, Aerial-NeRF enjoys the rendering of arbitrarily large scenes using a single GPU.
Meanwhile, we introduce a novel sampling strategy for aerial scenes, which enables to cover buildings by the sampling ranges from cameras at different heights.
In a broader comparison against SOTA models, our approach is substantially more efficient (only 1/4 used sampling points and 2 GB GPU memory saving) and 
compares favourably in terms of multiple commonly-used metrics.
Finally, we present SCUTic, a novel aerial dataset for large-scale university campus scenes with uneven camera trajectory, 
which can verify the robustness of rendering methods.

\bibliographystyle{IEEEtran}
\bibliography{IEEEabrv,ref}












\vspace{-10 mm}

\begin{IEEEbiography}
  [{\includegraphics[width=1in,height=1.25in,clip,keepaspectratio]{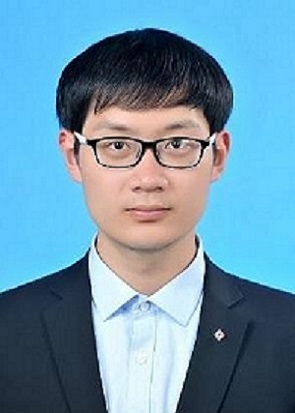}}]{Xiaohan Zhang}                                            
  received the M.Sc. degree with the School of Mathematics and
  Statistics, Shandong University, Weihai, China and received the B.S. degree from Xinjiang University, Urumqi, China. He is currently pursuing a doctor degree at 
  the School of Future Technology, South China University of Technology. His major is Electronic Information. His research interest is 3D vision, 
  including multi-view stereo, NeRF, GS and the application of 3D reconstruction technology in large-scale scenes. 
\end{IEEEbiography}  

\vspace{-10 mm}

\begin{IEEEbiography}
  [{\includegraphics[width=1in,height=1.25in,clip,keepaspectratio]{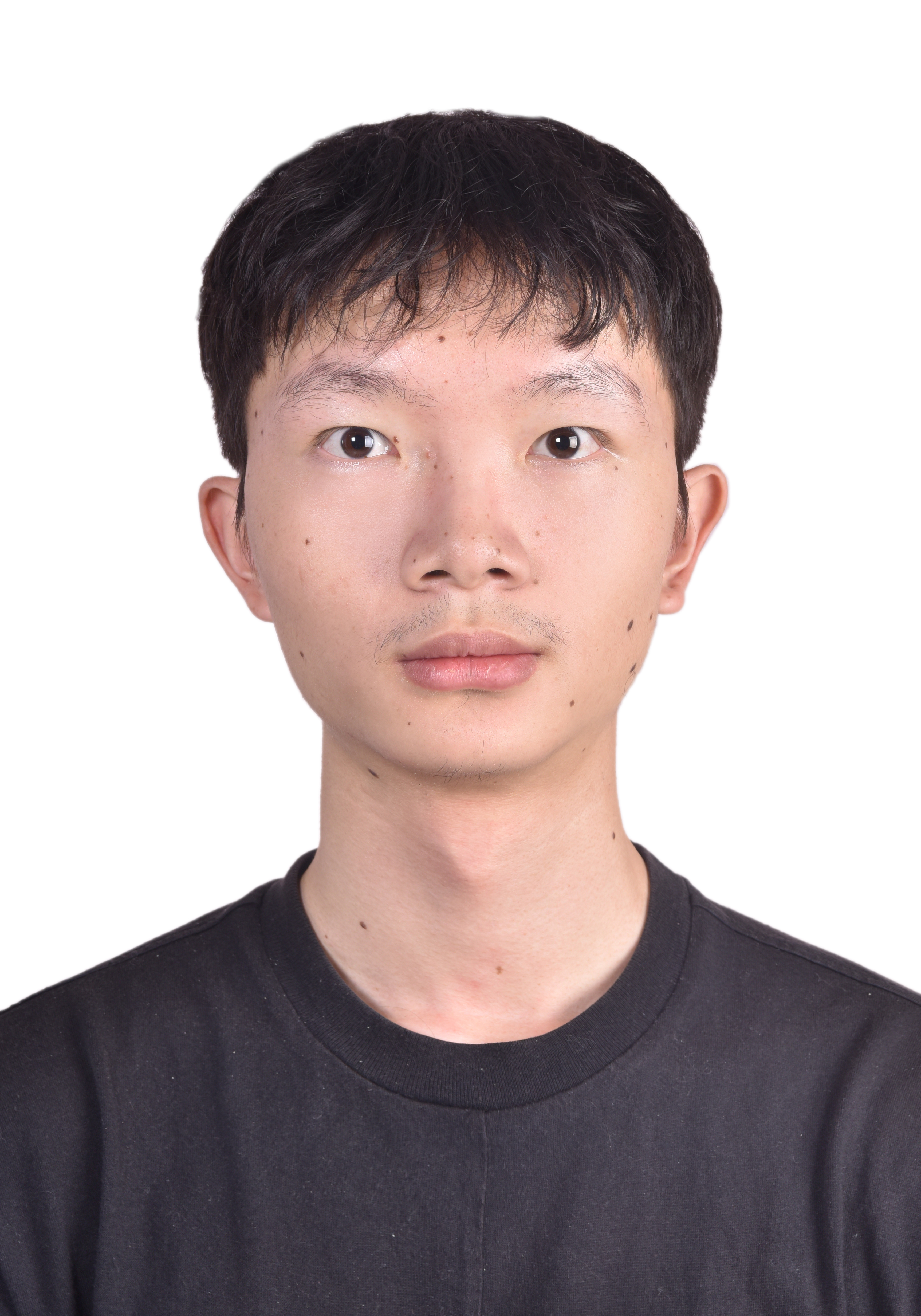}}]{Yukui Qiu} 
  is currently studying for a bachelor's degree at South China University of Technology. His main research interest is computer vision.
\end{IEEEbiography}  


\begin{IEEEbiography}
  [{\includegraphics[width=1in,height=1.25in,clip,keepaspectratio]{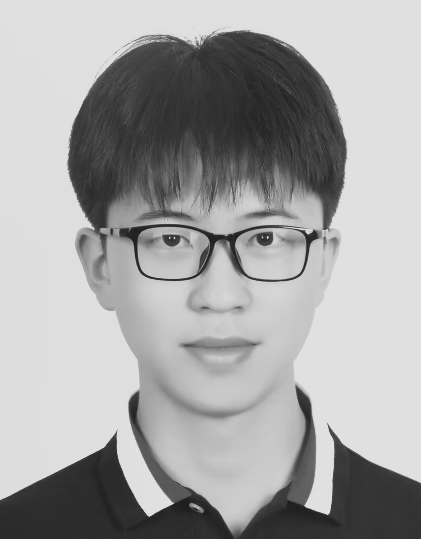}}]{Zhenyu Sun} 
  is currently working toward the bachelor's degree with South China University of Technology. His research interests include computer vision.
\end{IEEEbiography}  


\begin{IEEEbiography}
  [{\includegraphics[width=1in,height=1.25in,clip,keepaspectratio]{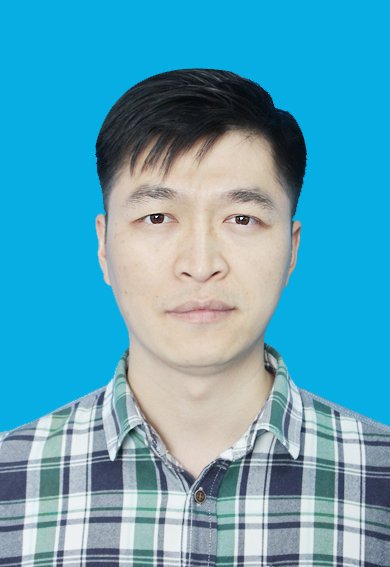}}]{Qi Liu} 
  is currently a Professor with the School of Future Technology at South China University of Technology. Dr. Liu received the Ph.D degree in Electrical Engineering from City University of Hong Kong, Hong Kong, China, in 2019. During 2018 - 2019, he was a Visiting Scholar at University of California Davis, CA, USA. From 2019 to 2022, he worked as a Research Fellow in the Department of Electrical and Computer Engineering, National University of Singapore, Singapore. His research interests include human-object interaction, AIGC, 3D scene reconstruction, and affective computing, etc. Dr. Liu has been an Associate Editor of the IEEE Systems Journal (2022-), and Digital Signal Processing (2022-). He was also Guest Editor for the IEEE Internet of Things Journal, IET Signal Processing, etc. He was the recipient of the Best Paper Award of IEEE ICSIDP in 2019.
\end{IEEEbiography}  

\end{document}